\documentclass[11pt,a4paper]{article}
\usepackage[hyperref]{acl2018}
\usepackage{times}
\usepackage{latexsym}

\usepackage{url}

\usepackage{graphicx}
\usepackage{caption}
\usepackage{booktabs}
\usepackage{enumitem}
\usepackage{xspace}
\usepackage{multirow}
\usepackage{algorithm}
\usepackage{algorithmic}

\usepackage{amsmath}
\usepackage{amsfonts}
\usepackage{amsthm}
\usepackage[makeroom]{cancel}

\usepackage{slashbox}
\usepackage{subcaption}
\usepackage{chngpage}

\aclfinalcopy 

\newcommand{\eos}{\texttt{\small eos}}

\DeclareMathOperator*{\E}{\mathbb{E}}

\newcommand{\EXP}[1]{\exp\left({#1}\right)}
\newcommand{\KL}[2]{\mathrm{KL}\left({#1} \| {#2} \right)}
\newcommand{\CE}[2]{\mathrm{CE}\left({#1} \| {#2} \right)}

\newcommand{\mb}[1]{\boldsymbol{\mathbf{#1}}}

\newtheorem{innercustomprop}{Proposition}
\newenvironment{customprop}[1]
  {\renewcommand\theinnercustomprop{#1}\innercustomprop}
  {\endinnercustomprop}
  
\newtheorem{innercustomcoro}{Corollary}
\newenvironment{customcoro}[1]
  {\renewcommand\theinnercustomcoro{#1}\innercustomcoro}
  {\endinnercustomcoro}

\newtheorem{prop}{Proposition}
\newtheorem{coro}{Corollary}

\allowdisplaybreaks

\newcommand*\samethanks[1][\value{footnote}]{\footnotemark[#1]}

\title{From Credit Assignment to Entropy Regularization: \\
	Two New Algorithms for Neural Sequence Prediction}

\author{Zihang Dai\thanks{~~~Equal contribution.}~~, Qizhe Xie\samethanks~~, Eduard Hovy \\
  Language Technologies Institute\\
  Carnegie Mellon University\\
  {\tt \{dzihang, qizhex, hovy\}@cs.cmu.edu}}

\date{}

\begin{document}
\maketitle
\begin{abstract}
In this work, we study the credit assignment problem in reward augmented maximum likelihood (RAML) learning, and establish a theoretical equivalence between the token-level counterpart of RAML and the entropy regularized reinforcement learning.
Inspired by the connection, we propose two sequence prediction algorithms, one extending RAML with fine-grained credit assignment and the other improving Actor-Critic with a systematic entropy regularization.
On two benchmark datasets, we show the proposed algorithms outperform RAML and Actor-Critic respectively, providing new alternatives to sequence prediction.
\end{abstract}

\section{Introduction}
\label{sec:intro}

Modeling and predicting discrete sequences is the central problem to many natural language processing tasks. 
In the last few years, the adaption of recurrent neural networks (RNNs) and the sequence-to-sequence model (seq2seq)~\citep{sutskever2014sequence,bahdanau2014neural} has led to a wide range of successes in conditional sequence prediction, including machine translation~\citep{sutskever2014sequence,bahdanau2014neural}, automatic summarization~\cite{rush2015neural}, image captioning~\citep{karpathy2015deep,vinyals2015show,xu2015show} and speech recognition~\cite{chan2016listen}.

Despite the distinct evaluation metrics for the aforementioned tasks, the standard training algorithm has been the same for all of them. 
Specifically, the algorithm is based on maximum likelihood estimation (MLE), which maximizes the log-likelihood of the ``ground-truth'' sequences empirically observed.\footnote{In this work, we use the terms ``ground-truth'' and ``reference'' to refer to the empirical observations interchangeably. } 

While largely effective, the MLE algorithm has two obvious weaknesses.
Firstly, the MLE training ignores the information of the task specific metric. 
As a result, the potentially large discrepancy between the log-likelihood during training and the task evaluation metric at test time can lead to a suboptimal solution. 
Secondly, MLE can suffer from the exposure bias, which refers to the phenomenon that the model is never exposed to its own failures during training, and thus cannot recover from an error at test time.
Fundamentally, this issue roots from the difficulty in statistically modeling the exponentially large space of sequences, where most combinations cannot be covered by the observed data.

To tackle these two weaknesses, there have been various efforts recently, which we summarize into two broad categories:
\begin{itemize}[leftmargin=1.0em,topsep=0.5em,itemsep=0em]
\item A widely explored idea is to directly optimize the task metric for sequences produced by the model, with the specific approaches ranging from minimum risk training (MRT)~\citep{shen2015minimum} and learning as search optimization (LaSO)~\citep{daume2005learning,wiseman2016sequence} to reinforcement learning (RL)~\citep{ranzato2015sequence,bahdanau2016actor}.
In spite of the technical differences, the key component to make these training algorithms \textit{practically efficient} is often a delicate credit assignment scheme, which transforms the sequence-level signal into dedicated smaller units (e.g., token-level or chunk-level), and allocates them to specific decisions, allowing for efficient optimization with a much lower variance.
For instance, the beam search optimization (BSO)~\citep{wiseman2016sequence} utilizes the position of margin violations to produce signals to the specific chunks, while the actor-critic (AC) algorithm~\citep{bahdanau2016actor} trains a critic to enable token-level signals.

\item Another alternative idea is to construct a task metric dependent target distribution, and train the model to match this task-specific target instead of the empirical data distribution.
As a typical example, the reward augmented maximum likelihood (RAML)~\citep{norouzi2016reward} defines the target distribution as the exponentiated pay-off (sequence-level reward) distribution.
This way, RAML not only can incorporate the task metric information into training, but it can also alleviate the exposure bias by exposing imperfect outputs to the model.
However, RAML only works on the sequence-level training signal.
\end{itemize}

In this work, we are intrigued by the question whether it is possible to incorporate the idea of fine-grained credit assignment into RAML. 
More specifically, inspired by the token-level signal used in AC, we aim to find the token-level counterpart of the sequence-level RAML, i.e., defining a token-level target distribution for each auto-regressive conditional factor to match.
Motived by the question, we first formally define the desiderata the token-level counterpart needs to satisfy and derive the corresponding solution (\S\ref{sec:theory}).
Then, we establish a theoretical connection between the derived token-level RAML and entropy regularized RL (\S\ref{sec:connection}).
Motivated by this connection, we propose two algorithms for neural sequence prediction, where one is the token-level extension to RAML, and the other a RAML-inspired improvement to the AC (\S\ref{sec:algo}).
We empirically evaluate the two proposed algorithms, and show different levels of improvement over the corresponding baseline.
We further study the importance of various techniques used in our experiments, providing practical suggestions to readers (\S\ref{sec:exp}).

\section{Token-level Equivalence of RAML}
\label{sec:theory}

We first introduce the notations used throughout the paper.
Firstly, capital letters will denote random variables and lower-case letters are the values to take.
As we mainly focus on conditional sequence prediction, we use $\mb{x}$ for the conditional input, and $\mb{y}$ for the target sequence.
With $\mb{y}$ denoting a sequence, $\mb{y}_{i}^{j}$ then denotes the subsequence from position $i$ to $j$ inclusively, while $y_t$ denotes the single value at position $t$.
Also, we use $|\mb{y}|$ to indicate the length of the sequence.
To emphasize the ground-truth data used for training, we add superscript $^*$ to the input and target, i.e., $\mb{x}^*$ and $\mb{y}^*$.
In addition, we use $\mathcal{Y}$ to denote the set of all possible sequences with one and only one $\eos$ symbol at the end, and $\mathcal{W}$ to denote the set of all possible symbols in a position.
Finally, we assume length of sequences in $\mathcal{Y}$ is bounded by $T$.

\subsection{Background: RAML}
\label{sec:raml}
As discussed in \S\ref{sec:intro}, given a ground-truth pair $(\mb{x}^*, \mb{y}^*)$, RAML defines the target distribution using the exponentiated pay-off of sequences, i.e.,
\begin{equation}\fontsize{9.5}{11.5}
\label{eqn:raml_target}
P_R(\mb{y} \mid \mb{x}^*, \mb{y}^*) = \frac{\EXP{R(\mb{y}; \mb{y}^*) / \tau}}{\sum_{\mb{y}' \in \mathcal{Y}} \EXP{R(\mb{y}'; \mb{y}^*) / \tau}},
\end{equation}
where $R(\mb{y}; \mb{y}^*)$ is the sequence-level reward, such as BLEU score, and $\tau$ is the temperature hyper-parameter controlling the sharpness.
With the definition, the RAML algorithm simply minimizes the cross entropy (CE) between the target distribution and the model distribution $P_\theta(\mb{Y} \mid \mb{x}^*)$, i.e., 
\begin{equation}\fontsize{9.5}{11.5}
\label{eqn:raml_objective}
\min_{\theta} \CE{P_R(\mb{Y} \mid \mb{x}^*, \mb{y}^*)}{P_\theta(\mb{Y} \mid \mb{x}^*)}.
\end{equation}
Note that, this is quite similar to the MLE training, except that the target distribution is different.
With the particular choice of target distribution, RAML not only makes sure the ground-truth reference remains the mode, but also allows the model to explore sequences that are not exactly the same as the reference but have relatively high rewards.

Compared to algorithms trying to directly optimize task metric, RAML avoids the difficulty of tracking and sampling from the model distribution that is consistently changing. Hence, RAML enjoys a much more stable optimization without the need of pretraining. 
However, in order to optimize the RAML objective (Eqn. \eqref{eqn:raml_objective}), one needs to sample from the exponentiated pay-off distribution, which is quite challenging in practice. Thus, importance sampling is often used~\citep{norouzi2016reward,ma2017softmax}. We leave the details of the practical implementation to Appendix \ref{sec:A-raml}.

\subsection{Token-level Target Distribution}
\label{sec:vaml_target}
Despite the appealing properties, RAML only operates on the sequence-level reward. 
As a result, the reward gap between any two sequences cannot be attributed to the responsible decisions precisely, which often leads to a low sample efficiency.
Ideally, since we rely on the auto-regressive factorization
$P_\theta(\mb{y} \mid \mb{x}^*) = \prod_{t=1}^{|\mb{y}|} P_\theta(y_t \mid \mb{y}_{1}^{t-1}, \mb{x}^*)$, 
the optimization would be much more efficient if we have the target distribution for each token-level factor $P_\theta(Y_t \mid \mb{y}_{1}^{t-1}, \mb{x}^*)$ to match.
Conceptually, this is exactly how the AC algorithm improves upon the vanilla sequence-level REINFORCE algorithm~\cite{ranzato2015sequence}.

With this idea in mind, we set out to find such a token-level target. Firstly, we assume the token-level target shares the form of a Boltzmann distribution but parameterized by some unknown negative energy function $Q_R$, i.e.,\footnote{To avoid clutter, the conditioning on $\mb{x}^*$ will be omitted in the sequel, assuming it's clear from the context.}
\begin{equation}\fontsize{9.5}{11.5}
\label{eqn:vaml_target}
P_{Q_R}(y_t \mid \mb{y}_{1}^{t-1}, \mb{y}^*)
= \frac{ \EXP{ Q_R(\mb{y}_{1}^{t-1}, y_t; \mb{y}^*) / \tau } }{ \sum_{w \in \mathcal{W}} \EXP{ Q_R(\mb{y}_{1}^{t-1}, w; \mb{y}^*) / \tau } }.
\end{equation}
Intuitively, $Q_R(\mb{y}_{1}^{t-1}, w; \mb{y}^*)$ measures how much \textit{future} pay-off one can expect if $w$ is generated, given the current status $\mb{y}_{1}^{t-1}$ and the reference $\mb{y}^*$. This quantity highly resembles the action-value function ($Q$-function) in reinforcement learning. As we will show later, it is indeed the case.

Before we state the desiderata for $Q_R$, we need to extend the definition of $R$ in order to evaluate the goodness of an unfinished partial prediction, i.e., sequences without an $\eos$ suffix. 
Let $\mathcal{Y}^-$ be the set of unfinished sequences, following \citet{bahdanau2016actor}, we define the pay-off function $R$ for a partial sequence $\hat{\mb{y}} \in \mathcal{Y}^-, |\hat{\mb{y}}| < T$ as 
\begin{equation}\fontsize{9.5}{11.5}
R(\hat{\mb{y}}; \mb{y}^*) = R(\hat{\mb{y}} + \eos; \mb{y}^*), 
\end{equation}
where the $+$ indicates string concatenation.

With the extension, we are ready to state two requirements for $Q_R$:
\begin{enumerate}[leftmargin=1.0em,topsep=0.5em,itemsep=0em]
\item \textbf{Marginal match}: 
	For $P_{Q_R}$ to be the token-level equivalence of $P_R$, the sequence-level marginal distribution induced by $P_{Q_R}$ must match $P_R$, i.e., for any $\mb{y} \in \mathcal{Y}$,
	\begin{equation}\fontsize{9.5}{11.5}
	\label{eqn:marginal_match}
	\prod_{t=1}^{|\mb{y}|} P_{Q_R}(y_t \mid \mb{y}_{1}^{t-1}) = P_R(\mb{y}).
	\end{equation}
	Note that there are infinitely many $Q_R$'s satisfying Eqn. \eqref{eqn:marginal_match}, because adding any constant value to $Q_R$ does not change the Boltzmann distribution, known as shift-invariance w.r.t. the energy.
\item \textbf{Terminal condition}: 
	Secondly, let's consider the value of $Q_R$ when emitting an $\eos$ symbol to immediately terminate the generation.
	As mentioned earlier, $Q_R$ measures the expected future pay-off. 
	Since the emission of $\eos$ ends the generation, the future pay-off can only come from the immediate increase of the pay-off.
	Thus, we require $Q_R$ to be the incremental pay-off when producing $\eos$, i.e.
	\begin{equation}\fontsize{9.5}{11.5}
	\label{eqn:terminal_condition}
	Q_R(\hat{\mb{y}}, \eos; \mb{y}^*) = R(\hat{\mb{y}} + \eos; \mb{y}^*) - R(\hat{\mb{y}}; \mb{y}^*),
	\end{equation}
	for any $\hat{\mb{y}} \in \mathcal{Y}^-$.
	Since Eqn. \eqref{eqn:terminal_condition} enforces the absolute of $Q_R$ at a point, it also solves the ambiguity caused by the shift-invariance property.
\end{enumerate}
Based on the two requirements, we can derive the form $Q_R$, which is summarized by Proposition \ref{thm:prop}.
\begin{prop}
\label{thm:prop}
	$P_{Q_R}$ and $Q_R$ satisfy requirements \eqref{eqn:marginal_match} and \eqref{eqn:terminal_condition} if and only if for any ground-truth pair $(\mb{x}^*, \mb{y}^*)$ and any sequence prediction $\mb{y} \in \mathcal{Y}$,
	\begin{align}\fontsize{9.5}{11.5}
	\label{eqn:recursion}
	Q_R(\mb{y}_{1}^{t-1}, y_t; \mb{y}^*) 
	= R(\mb{y}_{1}^t; \mb{y}^*)  - R(\mb{y}_{1}^{t-1}; \mb{y}^*) \nonumber\\\fontsize{9.5}{11.5}
	+\; \tau \log \sum_{w \in \mathcal{W}} \EXP{Q_R(\mb{y}_{1}^{t}, w; \mb{y}^*) / \tau},
	\end{align}
	when $t < |\mb{y}|$, and otherwise, i.e., when $t = |\mb{y}|$
	\begin{align}\fontsize{9.5}{11.5}
	\label{eqn:recursion_terminal}
	Q_R(\mb{y}_{1}^{t-1}, y_t; \mb{y}^*) 
	= R(\mb{y}_{1}^t; \mb{y}^*)  - R(\mb{y}_{1}^{t-1}; \mb{y}^*).
	\end{align}
\end{prop}
\begin{proof}
	See Appendix \ref{sec:A-main-proof}.
\end{proof}
Note that, instead of giving an explicit form for the token-level target distribution, Proposition \ref{thm:prop} only provides an equivalent condition in the form of an implicit recursion.
Thus, we haven't obtained a practical algorithm yet.
However, as we will discuss next, the recursion has a deep connection to entropy regularized RL, which ultimately inspires our proposed algorithms.

\section{Connection to Entropy-regularized RL}
\label{sec:connection}

Before we dive into the connection, we first give a brief review of the entropy-regularized RL.
For an in-depth treatment, we refer readers to \citep{ziebart2010modeling,schulman2017equivalence}.

\subsection{Background: Entropy-regularized RL}
Following the standard convention of RL, we denote a Markov decision process (MDP) by a tuple $\mathcal{M} = (\mathcal{S}, \mathcal{A}, p_{s}, r, \gamma)$, where $\mathcal{S}, \mathcal{A}, p_{s}, r, \gamma$ are the state space, action space, transition probability, reward function and discounting factor respectively.\footnote{In sequence prediction, we are only interested in the periodic (finite horizon) case. }

Based on the notation, the goal of entropy-regularized RL augments is to learn a policy $\pi(a_t \mid s_t)$ which maximizes the discounted expected future return and causal entropy~\cite{ziebart2010modeling}, i.e.,
{\fontsize{9.5}{11.5} \[ \max_{\pi} \sum_{t} \E_{s_t \sim \rho_s, a_t \sim \pi(\cdot\mid s_t)} \gamma^{t-1} [ r(s_t, a_t) + \alpha \mathcal{H}(\pi(\cdot \mid s_t))], \]}
where $\mathcal{H}$ denotes the entropy and $\alpha$ is a hyper-parameter controlling the relative importance between the reward and the entropy.
Intuitively, compared to standard RL, the extra entropy term encourages exploration and promotes multi-modal behaviors. 
Such properties are highly favorable in a complex environment.

Given an entropy-regularized MDP, for any fixed policy $\pi$, the state-value function $V^\pi(s)$ and the action-value function $Q^\pi$ can be defined as
{\fontsize{9.5}{11.5}
\begin{equation}
\label{eqn:fixed_policy_iteration}
\begin{aligned}
	V^\pi(s) &= \E_{a \sim \pi(\cdot \mid s)} [Q^\pi(s, a)] + \alpha \mathcal{H}(\pi(\cdot \mid s)), \\
	Q^\pi(s, a) &= r(s, a) + \E_{s' \sim \rho_s} [ \gamma V^\pi(s') ].
\end{aligned}
\end{equation}}

With the definitions above, it can further be proved~\citep{ziebart2010modeling,schulman2017equivalence} that the optimal state-value function $V^*$, the action-value function $Q^*$ and the corresponding optimal policy $\pi^*$ satisfy the following equations
{\fontsize{9.5}{11.5}
\begin{align}
\label{eqn:soft_V}
V^*(s) &= \alpha \log \sum_{a \in \mathcal{A}} \EXP{ Q^*(s, a) / \alpha }, \\
\label{eqn:soft_Q}
Q^*(s, a) &= r(s, a) + \gamma \E_{s' \sim \rho_s} [V^*(s')], \\
\pi^*(a \mid s) &= \frac{ \EXP{ Q^*(s, a) / \alpha } }{ \sum_{a' \in \mathcal{A}} \EXP{ Q^*(s, a') / \alpha } }.
\end{align}}
Here, Eqn. \eqref{eqn:soft_V} and \eqref{eqn:soft_Q} are essentially the entropy-regularized counterparts of the optimal Bellman equations in standard RL.
Following previous literature, we will refer to Eqn. \eqref{eqn:soft_V} and \eqref{eqn:soft_Q} as the optimal \textit{soft} Bellman equations, and the $V^*$ and $Q^*$ as optimal \textit{soft} value functions.

\subsection{An RL Equivalence of the Token-level RAML}
\label{sec:rl_equivalence}
To reveal the connection, it is convenient to define the incremental pay-off
\begin{equation}\fontsize{9.5}{11.5}
\label{eqn:reward}
r(\mb{y}_{1}^{t-1}, y_t; \mb{y}^*) = R(\mb{y}_{1}^t; \mb{y}^*)  - R(\mb{y}_{1}^{t-1}; \mb{y}^*),
\end{equation}
and the last term of Eqn. \eqref{eqn:recursion} as 
\begin{equation}\fontsize{9.5}{11.5}
\label{eqn:optimal_V}
V_R(\mb{y}_{1}^{t}; \mb{y}^*) = \tau \log \sum_{w \in \mathcal{W}} \EXP{Q_R(\mb{y}_{1}^{t}, w; \mb{y}^*) / \tau}
\end{equation}
Substituting the two definitions into Eqn. \eqref{eqn:recursion}, the recursion simplifies as
\begin{equation}\fontsize{9.5}{11.5}
\label{eqn:optimal_Q}
Q_R(\mb{y}_{1}^{t-1}, y_t; \mb{y}^*) =
r(\mb{y}_{1}^{t-1}, y_t; \mb{y}^*) + V_R(\mb{y}_{1}^{t}; \mb{y}^*).
\end{equation}
Now, it is easy to see that the Eqn. \eqref{eqn:optimal_V} and \eqref{eqn:optimal_Q}, which are derived from the token-level RAML, highly resemble the optimal soft Bellman equations \eqref{eqn:soft_V} and \eqref{eqn:soft_Q} in entropy-regularized RL. 
The following Corollary formalizes the connection.
\begin{coro}\label{thm:coro}
	For any ground-truth pair $(\mb{x}^*, \mb{y}^*)$, the recursion specified by Eqn. \eqref{eqn:reward}, \eqref{eqn:optimal_V} and \eqref{eqn:optimal_Q} is equivalent to the optimal soft Bellman equation of a ``deterministic'' MDP in entropy-regularized reinforcement learning, denoted as $\mathcal{M}_R$, where 
	\begin{itemize}[topsep=0.5em,itemsep=0em]
		\item the state space $\mathcal{S}$ corresponds to $\mathcal{Y}^-$,
		\item the action space $\mathcal{A}$ corresponds to $\mathcal{W}$,
		\item the transition probability $\rho_s$ is a deterministic process defined by string concatenation
		\item the reward function $r$ corresponds to the incremental pay-off defined in Eqn. \eqref{eqn:reward},
		\item the discounting factor $\gamma = 1$,
		\item the entropy hyper-parameter $\alpha = \tau$,
		\item and a period terminates either when $\eos$ is emitted or when its length reaches $T$ and we enforce the generation of $\eos$.
	\end{itemize}
	Moreover, the optimal soft value functions $V^*$ and $Q^*$ of the MDP exactly match the $V_R$ and $Q_R$ defined by Eqn. \eqref{eqn:optimal_V} and \eqref{eqn:optimal_Q} respectively.
	The optimal policy $\pi^*$ is hence equivalent to the token-level target distribution $P_{Q_R}$. 
\end{coro}
\begin{proof}
	See Appendix \ref{sec:A-main-proof}.
\end{proof}
The connection established by Corollary \ref{thm:coro} is quite inspiring:
\begin{itemize}[leftmargin=1.0em,topsep=0.5em,itemsep=0em]
\item Firstly, it provides a rigorous and generalized view of the connection between RAML and entropy-regularized RL.
In the original work, \citet{norouzi2016reward} point out RAML can be seen as reversing the direction of  $\KL{P_\theta}{P_R}$, which is a sequence-level view of the connection.
Now, with the equivalence between the token-level target $P_{Q_R}$ and the optimal $Q^*$, it generalizes to matching the future action values consisting of both the reward and the entropy.
\item Secondly, due to the equivalence, if we solve the optimal soft $Q$-function of the corresponding MDP, we directly obtain the token-level target distribution. This hints at a practical algorithm with token-level credit assignment.
\item Moreover, since RAML is able to improve upon MLE by injecting entropy, the entropy-regularized RL counterpart of the standard AC algorithm should also lead to an improvement in a similar manner.
\end{itemize}

\section{Proposed Algorithms}
\label{sec:algo}
In this section, we explore the insights gained from Corollary \ref{thm:coro} and present two new algorithms for sequence prediction.

\subsection{Value Augmented Maximum Likelihood}
\label{sec:vaml}
The first algorithm we consider is the token-level extension of RAML, which we have been discussing since \S\ref{sec:theory}.
As mentioned at the end of \S\ref{sec:vaml_target}, Proposition \ref{thm:prop} only gives an implicit form of $Q_R$, and so is the token-level target distribution $P_{Q_R}$ (Eqn. \eqref{eqn:vaml_target}).
However, thanks to Corollary \ref{thm:coro}, we now know that $Q_R$ is the same as the optimal soft action-value function $Q^*$ of the entropy-regularized MDP $\mathcal{M}_R$. 
Hence, by finding the $Q^*$, we will have access to $P_{Q_R}$.

At the first sight, it seems recovering $Q^*$ is as difficult as solving the original sequence prediction problem, because solving $Q^*$ from the MDP is essentially the same as learning the optimal policy for sequence prediction.
However, it is not true because $Q_R$ (i.e., $P_{Q_R}$) can condition on the correct reference $\mb{y}^*$. In contrast, the model distribution $P_\theta$ can only depend on $\mb{x}^*$.
Therefore, the function approximator trained to recover $Q^*$ can take $\mb{y}^*$ as input, making the estimation task much easier.
Intuitively, when recovering $Q^*$, we are trying to train an ideal ``oracle'', which has access to the ground-truth reference output, to decide the best behavior (policy) given any arbitrary (good or not) state.

Thus, following the reasoning above, we first train a parametric function approximator $Q_\phi$ to search the optimal soft action value.
In this work, for simplicity, we employ the Soft Q-learning algorithm~\citep{schulman2017equivalence} to perform the policy \textit{optimization}.
In a nutshell, Soft Q-Learning is the entropy-regularized version of Q-Learning, an off-policy algorithm which minimizes the mean squared soft Bellman residual according to Eqn. \eqref{eqn:soft_Q}.
Specifically, given ground-truth pair $(\mb{x}^*, \mb{y}^*)$, for any trajectory $\mb{y} \in \mathcal{Y}$, the training objective is 
\begin{equation}\fontsize{9.5}{11.5}
\label{eqn:soft_q_learning}
\min_{\phi} \sum_{t=1}^{|\mb{y}|} \left[ Q_\phi(\mb{y}_{1}^{t-1}, y_t; \mb{y}^*) - \hat{Q}_\phi(\mb{y}_{1}^{t-1}, y_t; \mb{y}^*) \right]^2,
\end{equation}
where $\fontsize{9.5}{11.5} \hat{Q}_\phi(\mb{y}_{1}^{t-1}, y_t; \mb{y}^*) = r(\mb{y}_{1}^{t-1}, y_t; \mb{y}^*) + V_\phi(\mb{y}_{1}^{t}; \mb{y}^*)$ is the one-step look-ahead target Q-value, and $\fontsize{9.5}{11.5} V_\phi(\mb{y}_{1}^{t}; \mb{y}^*) = \tau \log \sum_{w \in \mathcal{W}} \EXP{ Q_\phi(\mb{y}_{1}^{t}, w; \mb{y}^*) / \tau}$ as defined in Eqn. \eqref{eqn:soft_V}.
In the recent instantiation of Q-Learning~\citep{mnih2015human}, to stabilize training, the target Q-value is often estimated by a separate slowly updated target network.
In our case, as we have access to a significant amount of reference sequences, we find the target network not necessary. 
Thus, we directly optimize Eqn. \eqref{eqn:soft_q_learning} using gradient descent, and let the gradient flow through both $Q_\phi(\mb{y}_{1}^{t-1}, y_t; \mb{y}^*)$ and $V_\phi(\mb{y}_{1}^{t}; \mb{y}^*)$~\citep{baird1995residual}.

After the training of $Q_\phi$ converges, we fix the parameters of $Q_\phi$, and optimize the cross entropy $\CE{P_{Q_\phi}}{P_\theta}$ w.r.t. the model parameters $\theta$, which is equivalent to\footnote{See Appendix \ref{sec:A-other-proof} for a detailed derivation.}
\begin{equation} \fontsize{9}{11.5}
\label{eqn:vaml_objective}
\min_{\theta} \E_{\mb{y} \sim P_{Q_\phi}} \left[ \sum_{t=1}^{|\mb{y}|} \CE{P_{Q_\phi}(Y_t \mid \mb{y}_1^{t-1})}{ P_{\theta}(Y_t \mid \mb{y}_1^{t-1}) } \right].
\end{equation}
Compared to the of objective of RAML in Eqn. \eqref{eqn:raml_objective}, having access to $P_{Q_\phi}(Y_t \mid \mb{y}_1^{t-1})$ allows us to provide a distinct token-level target for each conditional factor $P_{\theta}(Y_t \mid \mb{y}_1^{t-1})$ of the model.
While directly sampling from $P_R$ is practically infeasible (\S\ref{sec:raml}), having a parametric target distribution $P_{Q_\phi}$ makes it theoretically possible to sample from $P_{Q_\phi}$ and perform the optimization.
However, empirically, we find the samples from $P_{Q_\phi}$ are not diverse enough  (\S\ref{sec:exp}). Hence, we fall back to the same importance sampling approach (see Appendix \ref{sec:A-vaml}) as used in RAML.

Finally, since the algorithm utilizes the optimal soft action-value function to construct the token-level target, we will refer to it as value augmented maximum likelihood (VAML) in the sequel.

\subsection{Entropy-regularized Actor Critic}
\label{sec:erac}
The second algorithm follows the discussion at the end of \S\ref{sec:rl_equivalence}, which is essentially an actor-critic algorithm based on the entropy-regularized MDP in Corollary \ref{thm:coro}. 
For this reason, we name the algorithm entropy-regularized actor critic (ERAC).
As with standard AC algorithm, the training process interleaves the evaluation of current policy using the parametric critic $Q_\phi$ and the optimization of the actor policy $\pi_\theta$ given the current critic.

\paragraph{Critic Training.}
The critic is trained to perform policy \textit{evaluation} using the temporal difference learning (TD), which minimizes the TD error 
\begin{equation}\fontsize{9.5}{11.5}
\label{eqn:erac_critic}
\min_{\phi} \E_{\mb{y} \sim \pi_{\theta}} \sum_{t=1}^{|\mb{y}|} \left[ Q_\phi(\mb{y}_{1}^{t-1}, y_t; \mb{y}^*) - \hat{Q}_{\bar{\phi}}(\mb{y}_{1}^{t-1}, y_t; \mb{y}^*) \right]^2
\end{equation}
where the TD target $\hat{Q}_{\bar{\phi}}$ is constructed based on fixed policy iteration in Eqn. \eqref{eqn:fixed_policy_iteration}, i.e.,
{\fontsize{9.5}{11.5}
\label{eqn:erac_critic_td}
\begin{align}
\hat{Q}_{\bar{\phi}}(\mb{y}_{1}^{t-1}, y_t; \mb{y}^*) 
&= r(\mb{y}_{1}^{t-1}, y_t) + \tau \, \mathcal{H}(\pi_\theta(\cdot \mid \mb{y}_1^t)) \nonumber\\
+& \sum_{w \in \mathcal{W}} \pi_\theta(w \mid \mb{y}_1^t) Q_{\bar{\phi}}(\mb{y}_1^t, w; \mb{y}^*).
\end{align}}
It is worthwhile to emphasize that the objective \eqref{eqn:erac_critic} trains the critic $Q_\phi$ to evaluate the current policy. 
Hence, it is entirely different from the objective \eqref{eqn:soft_q_learning}, which is performing policy optimization by Soft Q-Learning.
Also, the trajectories $\mb{y}$ used in \eqref{eqn:erac_critic} are sequences drawn from the actor policy $\pi_\theta$, while objective \eqref{eqn:soft_q_learning} theoretically accepts any trajectory since Soft Q-Learning can be fully off-policy.\footnote{Different from \citet{bahdanau2016actor}, we don't use a delayed actor network to collect trajectories for critic training.}
Finally, following \citet{bahdanau2016actor}, the TD target $\hat{Q}_{\bar{\phi}}$ in Eqn. \eqref{eqn:fixed_policy_iteration} is evaluated using a target network, which is indicated by the bar sign above the parameters, i.e., $\bar{\phi}$. 
The target network is slowly updated by linearly interpolating with the up-to-date network, i.e., the update is $\bar{\phi} \leftarrow \beta \phi + (1-\beta) \bar{\phi}$ for $\beta$ in $(0, 1)$ \citep{lillicrap2015continuous}.

We also adapt another technique proposed by \citet{bahdanau2016actor}, which smooths the critic by minimizing the ``variance'' of Q-values, i.e., 
\begin{equation*}\fontsize{9}{11.5}
\min_{\phi} \lambda_\text{var} \E_{\mb{y} \sim \pi_{\theta}} \sum_{t=1}^{|\mb{y}|} \sum_{w \in \mathcal{W}} \left[ Q_\phi(\mb{y}_{1}^{t}, w; \mb{y}^*) - \bar{Q}_\phi(\mb{y}_{1}^{t}; \mb{y}^*) \right]^2
\end{equation*}
where $\fontsize{10}{12} \bar{Q}_\phi(\mb{y}_{1}^{t}; \mb{y}^*) = \frac{1}{|\mathcal{W}|} \sum_{w' \in \mathcal{W}} Q_\phi(\mb{y}_{1}^{t}, w'; \mb{y}^*)$ is the mean Q-value, and $\lambda_\text{var}$ is a hyper-parameter controlling the relative weight between the TD loss and the smooth loss.

\paragraph{Actor Training.} Given the critic $Q_\phi$, the actor gradient (to maximize the expected return) is given by the policy gradient theorem of the entropy-regularized RL~\citep{schulman2017equivalence}, which has the form
{\fontsize{9}{11.5}
\begin{align}
\label{eqn:erac_actor}
\E_{\mb{y} \sim \pi_{\theta}} \sum_{t=1}^{|\mb{y}|} & \sum_{w \in \mathcal{W}} \nabla_\theta \pi_{\theta}(w \mid \mb{y}_1^{t-1}) Q_\phi(\mb{y}_1^{t-1}, w; \mb{y}^*) \nonumber\\ 
&+ \tau \nabla_\theta \mathcal{H}(\pi_{\theta}(\cdot \mid \mb{y}_1^{t-1})).
\end{align}}
Here, for each step $t$, we follow \citet{bahdanau2016actor} to sum over the entire symbol set $\mathcal{W}$, instead of using the single sample estimation often seen in RL. Hence, no baseline is employed.
It is worth mentioning that Eqn. \eqref{eqn:erac_actor} is \textit{not} simply adding an entropy term to the standard policy gradient as in A3C~\citep{mnih2016asynchronous}.
The difference lies in that the critic $Q_\phi$ trained by Eqn. \eqref{eqn:erac_critic} additionally captures the \textit{entropy from future steps},  
while the $\nabla_\theta \mathcal{H}$ term only captures the entropy of the current step.

Finally, similar to \citep{bahdanau2016actor}, we find it necessary to first pretrain the actor using MLE and then pretrain the critic before the actor-critic training.
Also, to prevent divergence during actor-critic training, it is helpful to continue performing MLE training along with Eqn. \eqref{eqn:erac_actor}, though using a smaller weight $\lambda_\text{mle}$.

\section{Related Work}
\paragraph{Task Loss Optimization and Exposure Bias} Apart from the previously introduced RAML, BSO, Actor-Critic (\S\ref{sec:intro}), MIXER~\citep{ranzato2015sequence} also utilizes chunk-level signals where the length of chunk grows as training proceeds. 
In contrast, minimum risk training~\citep{shen2015minimum} directly optimizes sentence-level BLEU. As a result, it requires a large number (100) of samples per data to work well.
To solve the exposure bias, scheduled sampling~\citep{bengio2015scheduled} adopts a curriculum learning strategy to bridge the training and the inference. 
Professor forcing~\citep{lamb2016professor} introduces an adversarial training mechanism to encourage the dynamics of the model to be the same at training time and inference time.
For image caption, self-critic sequence training (SCST)~\citep{rennie2016self} extends the MIXER algorithm with an improved baseline based on the current model performance.

\paragraph{Entropy-regularized RL} 
Entropy regularization been explored by early work in RL and inverse RL~\citep{williams1991function,ziebart2008maximum}.
Lately, \citet{schulman2017equivalence} establish the equivalence between policy gradients and Soft Q-Learning under entropy-regularized RL.
Motivated by the multi-modal behavior induced by entropy-regularized RL, \citet{haarnoja2017reinforcement} apply energy-based policy and Soft Q-Learning to continuous domain.
Later, \citet{nachum2017bridging} proposes path consistency learning, which can be seen as a multi-step extension to Soft Q-Learning.
More recently, in the domain of simulated control, \citet{haarnoja2018soft} also consider the actor critic algorithm under the framework of entropy regularized reinforcement learning.
Despite the conceptual similarity to ERAC presented here, \citet{haarnoja2018soft} focuses on continuous control and employs the advantage actor critic variant as in~\cite{mnih2016asynchronous}, while ERAC follows the Q actor critic as in~\cite{bahdanau2016actor}.

\section{Experiments}
\label{sec:exp}

\begin{table*}[t]
	\begin{adjustwidth}{-.5in}{-.5in}
	\centering
	\begin{tabular}{l|ccc|ccc|ccc}
		\toprule
		& \multicolumn{3}{c|}{\bf MT (w/o input feeding)} & \multicolumn{3}{c|}{\bf MT (w/ input feeding)}  & \multicolumn{3}{c}{\bf Image Captioning} \\
		\bf Algorithm & Mean &  Min & Max & Mean & Min & Max & Mean & Min & Max \\
		\midrule
		MLE  & 27.01 $\pm$ 0.20 & 26.72 & 27.27 & 28.06 $\pm$ 0.15 & 27.84 & 28.22 & 29.54 $\pm$ 0.21 & 29.27 & 29.89 \\
		\midrule
		RAML & 27.74 $\pm$ 0.15 & 27.47 & 27.93 & 28.56 $\pm$ 0.15 & 28.35 & 28.80 & 29.84 $\pm$ 0.21 & 29.50 & 30.17 \\
		VAML & \textbf{28.16 $\pm$ 0.11} & \textbf{28.00} & \textbf{28.26} & \textbf{28.84 $\pm$ 0.10} & \textbf{28.62} & \textbf{28.94} & \textbf{29.93 $\pm$ 0.22} & \textbf{29.51} & \textbf{30.24} \\ 
		\midrule
		AC & 28.04 $\pm$ 0.05  & 27.97 & 28.10 & 29.05 $\pm$ 0.06  & 28.95 & 29.16 & 30.90 $\pm$ 0.20 & 30.49 & 31.16 \\
		ERAC & \textbf{28.30 $\pm$ 0.06}& \textbf{28.25} & \textbf{28.42} & \textbf{29.31 $\pm$ 0.04}& \textbf{29.26} & \textbf{29.36} & \textbf{31.44 $\pm$ 0.22} & \textbf{31.07} & \textbf{31.82} \\
		\bottomrule
	\end{tabular}
	\caption{Test results on two benchmark tasks. Bold faces highlight the best in the corresponding category.}
	\label{tab:result}
	\end{adjustwidth}
\end{table*}

\subsection{Experiment Settings}
\label{sec:exp-setting}
In this work, we focus on two sequence prediction tasks: machine translation and image captioning. 
Due to the space limit, we only present the information necessary to compare the empirical results at this moment.
For a more detailed description, we refer readers to Appendix \ref{sec:A-implementation} and the code\footnote{\scriptsize \url{https://github.com/zihangdai/ERAC-VAML}}.

\paragraph{Machine Translation} Following~\citet{ranzato2015sequence}, we evaluate on IWSLT 2014 German-to-English dataset~\citep{mauro2012wit3}. 
The corpus contains approximately $153K$ sentence pairs in the training set. 
We follow the pre-processing procedure used in~\citep{ranzato2015sequence}.

Architecture wise, we employ a seq2seq model with dot-product attention~\citep{bahdanau2014neural,luong2015effective}, where the encoder is a bidirectional LSTM~\citep{hochreiter1997long} with each direction being size $128$, and the decoder is another LSTM of size $256$.
Moreover, we consider two variants of the decoder, one using the input feeding technique~\citep{luong2015effective} and the other not.

For all algorithms, the sequence-level BLEU score is employed as the pay-off function $R$, while the corpus-level BLEU score~\citep{papineni2002bleu} is used for the final evaluation. 
The sequence-level BLEU score is scaled up by the sentence length so that the scale of the immediate reward at each step is invariant to the length. 

\paragraph{Image Captioning}
For image captioning, we consider the MSCOCO dataset~\citep{lin2014microsoft}. 
We adapt the same preprocessing procedure and the train/dev/test split used by~\citet{karpathy2015deep}.

The NIC~\citep{vinyals2015show} is employed as the baseline model, where a feature vector of the image is extracted by a pre-trained CNN and then used to initialize the LSTM decoder. 
Different from the original NIC model, we employ a pre-trained $101$-layer ResNet~\citep{he2016deep} rather than a GoogLeNet as the CNN encoder. 

For training, each image-caption pair is treated as an i.i.d. sample, and sequence-level BLEU score is used as the pay-off. For testing, the standard multi-reference BLEU4 is used.

\subsection{Comparison with the Direct Baseline}
Firstly, we compare ERAC and VAML with their corresponding direct baselines, namely AC~\citep{bahdanau2016actor} and RAML~\citep{norouzi2016reward} respectively.
As a reference, the performance of MLE is also provided.

Due to non-neglected performance variance observed across different runs, we run each algorithm for 9 times with different random seeds,\footnote{For AC, ERAC and VAML, 3 different critics are trained first, and each critic is then used to train 3 actors.} and report the average performance, the standard deviation and the performance range (min, max).

\paragraph{Machine Translation}
The results on MT are summarized in the left half of Tab. \ref{tab:result}. 
Firstly, all four advanced algorithms significantly outperform the MLE baseline.
More importantly, both VAML and ERAC improve upon their direct baselines, RAML and AC, by a clear margin on average.
The result suggests the two proposed algorithms both well combine the benefits of a delicate credit assignment scheme and the entropy regularization, achieving improved performance.

\paragraph{Image Captioning}
The results on image captioning are shown in the right half of Tab. \ref{tab:result}. 
Despite the similar overall trend, the improvement of VAML over RAML is smaller compared to that in MT.
Meanwhile, the improvement from AC to ERAC becomes larger in comparison.
We suspect this is due to the multi-reference nature of the MSCOCO dataset, where a larger entropy is preferred.
As a result, the explicit entropy regularization in ERAC becomes immediately fruitful. 
On the other hand, with multiple references, it can be more difficult to learn a good oracle $Q^*$ (Eqn. \eqref{eqn:optimal_Q}).
Hence, the token-level target can be less accurate, resulting in smaller improvement.

\subsection{Comparison with Existing Work}
To further evaluate the proposed algorithms, we compare ERAC and VAML with the large body of existing algorithms evaluated on IWSTL 2014.
As a note of caution, previous works don't employ the exactly same architectures (e.g. number of layers, hidden size, attention type, etc.).
Despite that, for VAML and ERAC, we use an architecture that is most similar to the majority of previous works, which is the one described in \S\ref{sec:exp-setting} with input feeding.

Based on the setting, the comparison is summarized in Table \ref{tab:comparison}.\footnote{For a more detailed comparison of performance together with the model architectures, see Table \ref{tab:comparison_detailed} in Appendix \ref{sec:A-comparison}.}
As we can see, both VAML and ERAC outperform previous methods, with ERAC leading the comparison with a significant margin.
This further verifies the effectiveness of the two proposed algorithms.
\begin{table}[!h]
	\centering
	\begin{tabular}{l|c}
		\toprule
		\bf Algorithm & \bf BLEU \\
		\midrule
		MIXER~\cite{ranzato2015sequence} & 20.73 \\
		BSO~\cite{wiseman2016sequence} & 27.9 \\
		Q(BLEU)~\cite{li2017learning} & 28.3 \\
		AC~\cite{bahdanau2016actor} & 28.53 \\
		RAML~\cite{ma2017softmax} & 28.77 \\
		\midrule
		VAML & 28.94 \\
		ERAC & \bf 29.36 \\
		\bottomrule
	\end{tabular}
	\caption{Comparison with existing algorithms on IWSTL 2014 dataset for MT. All numbers of previous algorithms are from the original work.}
	\label{tab:comparison}
\end{table}

\subsection{Ablation Study}
Due to the overall excellence of ERAC, we study the importance of various components of it, hopefully offering a practical guide for readers. 
As the input feeding technique largely slows down the training, we conduct the ablation based on the model variant \textit{without} input feeding.

\begin{table}[!h]
	\centering
	\begin{tabular}{l|cccc}
		\toprule
		\backslashbox{$\lambda_\text{var}$}{$\beta$}& 0.001 & 0.01 & 0.1 & 1 \\
		\midrule
		$0$ & 27.91 & 26.27$^\dagger$ & 28.88 & 27.38$^\dagger$ \\
		$0.001$ & \textbf{29.41} & 29.26 & 29.32 & 27.44 \\
		\bottomrule
	\end{tabular}
	\caption{\fontsize{10}{12} Average validation BLEU of ERAC. As a reference, the average BLEU is 28.1 for MLE. $\lambda_\text{var}=0$ means not using the smoothing technique. $\beta=1$ means not using a target network. $^\dagger$ indicates excluding extreme values due to divergence.}
	\label{tab:ab_tgtnet_smooth}
\end{table}
Firstly, we study the importance of two techniques aimed for training stability, namely the target network and the smoothing technique (\S\ref{sec:erac}). 
Based on the MT task, we vary the update speed $\beta$ of the target critic, and the $\lambda_\text{var}$, which controls the strength of the smoothness regularization. 
The average \textit{validation} performances of different hyper-parameter values are summarized in Tab. \ref{tab:ab_tgtnet_smooth}. 
\begin{itemize}[leftmargin=1.0em,topsep=0.5em,itemsep=0em]
\item Comparing the two rows of Tab. \ref{tab:ab_tgtnet_smooth}, the smoothing technique consistently leads to performance improvement across all values of $\tau$. 
In fact, removing the smoothing objective often causes the training to diverge, especially when $\beta=0.01$ and $1$.
But interestingly, we find the divergence does not happen if we update the target network a little bit faster ($\beta=0.1$) or quite slowly ($\beta=0.001$).
\item In addition, even with the smoothing technique, the target network is still necessary. 
When the target network is not used ($\beta = 1$), the performance drops below the MLE baseline. 
However, as long as a target network is employed to ensure the training stability, the specific choice of target network update rate does not matter as much.
Empirically, it seems using a slower ($\beta=0.001$) update rate yields the best result.
\end{itemize}

\begin{figure*}[!h]
	\centering
	\begin{subfigure}[t]{0.45\textwidth}
		\centering
		\includegraphics[scale=0.4]{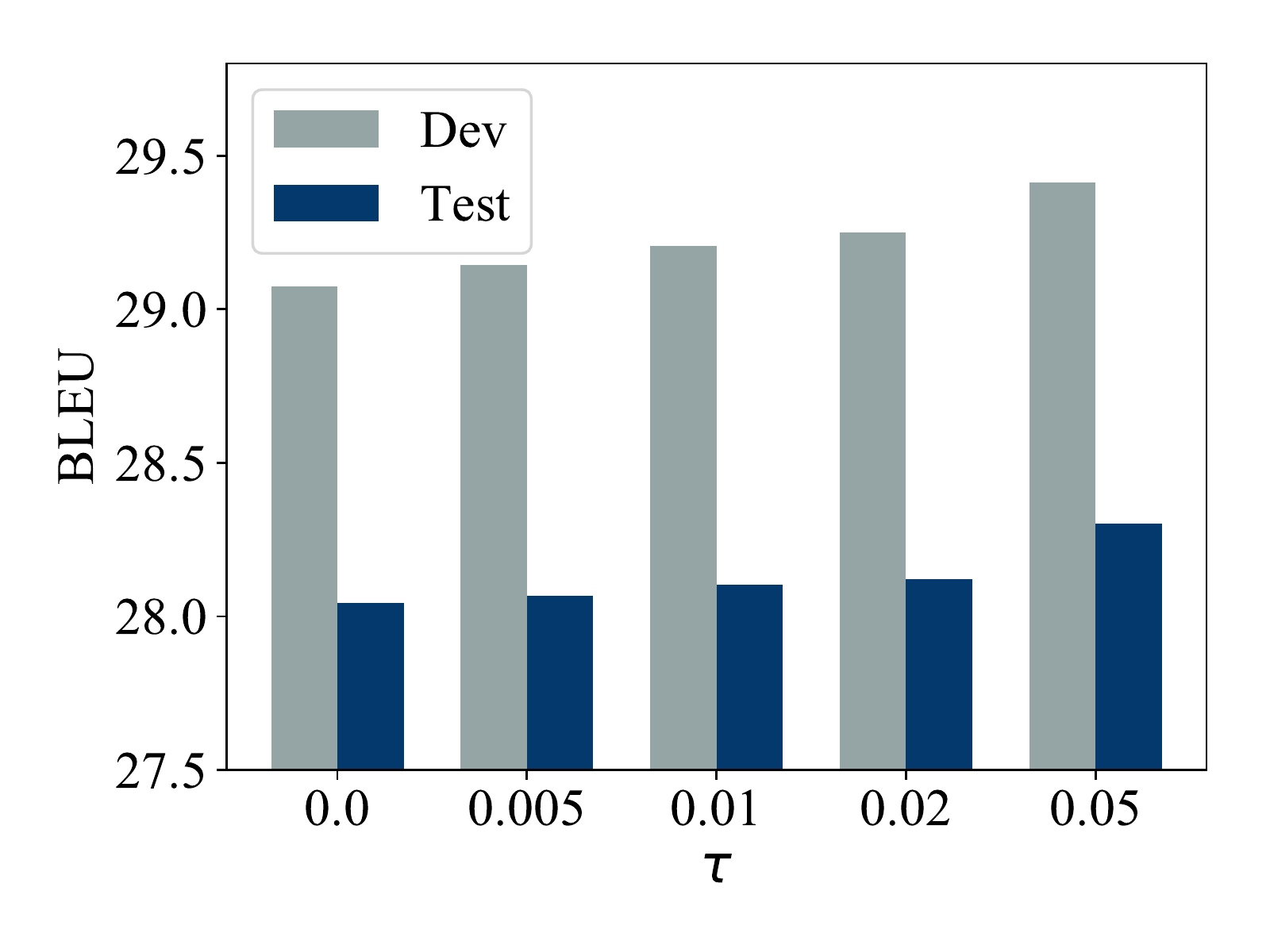}
		\caption{Machine translation}
	\end{subfigure}%
	~ 
	\begin{subfigure}[t]{0.45\textwidth}
		\centering
		\includegraphics[scale=0.4]{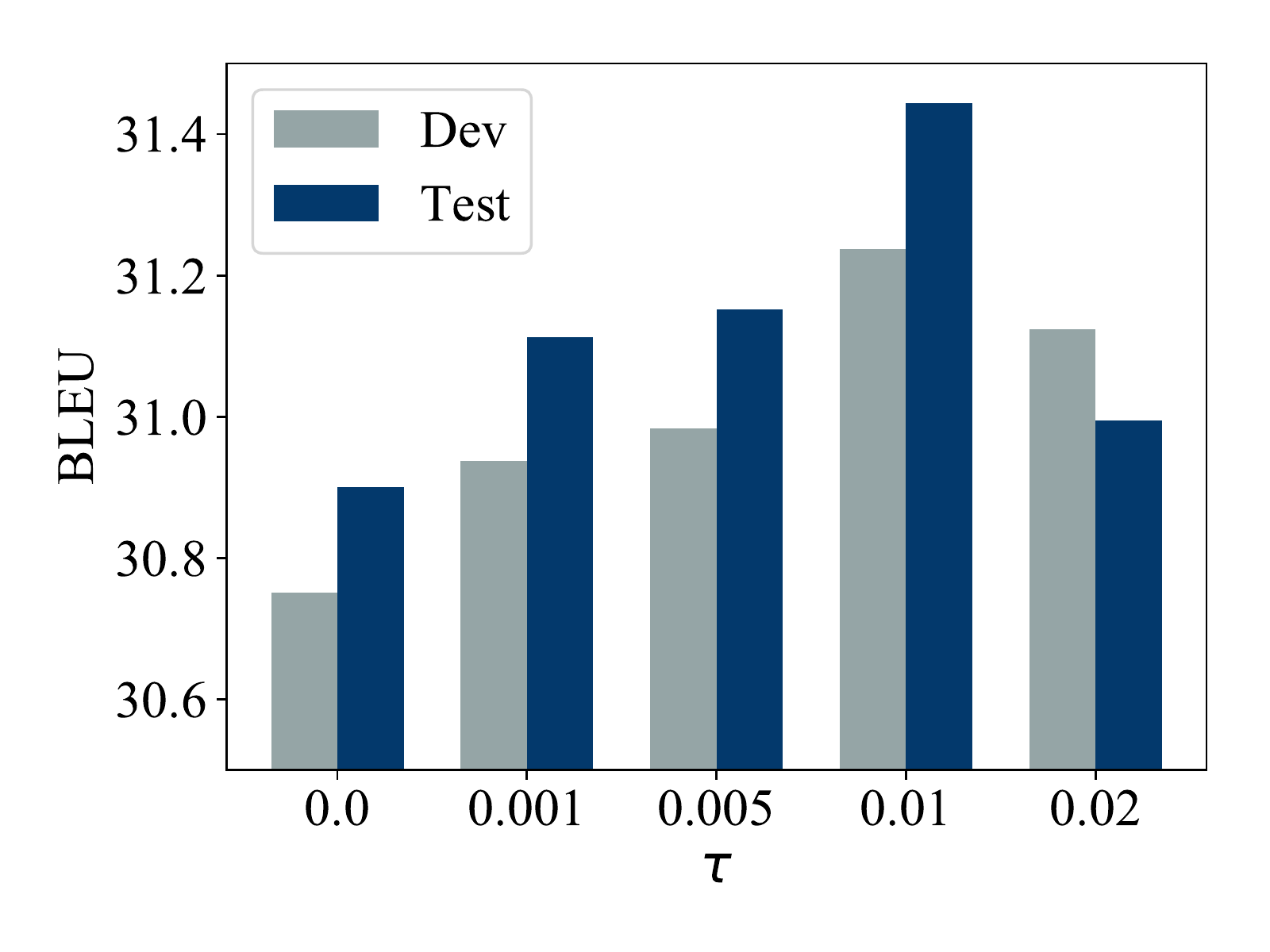}
		\caption{Image captioning}
	\end{subfigure}
	\caption{\fontsize{10}{12} ERAC's average performance over multiple runs on two tasks when varying $\tau$.}
	\label{fig:tau_ab}
\end{figure*}

Next, we investigate the effect of enforcing different levels of entropy by varying the entropy hyper-parameter $\tau$.
As shown in Fig. \ref{fig:tau_ab}, it seems there is always a sweet spot for the level of entropy.
On the one hand, posing an over strong entropy regularization can easily cause the actor to diverge. 
Specifically, the model diverges when $\tau$ reaches $0.03$ on the image captioning task or $0.06$ on the machine translation task. 
On the other hand, as we decrease $\tau$ from the best value to 0, the performance monotonically decreases as well. 
This observation further verifies the effectiveness of entropy regularization in ERAC, which well matches our theoretical analysis.

Finally, as discussed in \S\ref{sec:erac}, ERAC takes the effect of future entropy into consideration, and thus is different from simply adding an entropy term to the standard policy gradient as in A3C~\citep{mnih2016asynchronous}.
To verify the importance of explicitly modeling the entropy from future steps, we compared ERAC with the variant that only applies the entropy regularization to the actor but not to the critic.
In other words, the $\tau$ is set to 0 when performing policy evaluating according to Eqn. \eqref{eqn:erac_critic_td}, while the $\tau$ for the entropy gradient in Eqn. \eqref{eqn:erac_actor} remains.
The comparison result based on 9 runs on test set of IWSTL 2014 is shown in Table \ref{tab:future_ent}.
As we can see, simply adding a local entropy gradient does not even improve upon the AC.
This further verifies the difference between ERAC and A3C, and shows the importance of taking future entropy into consideration.
\begin{table}[!h]
	\centering
	\begin{tabular}{l|cc}
		\toprule
		\bf Algorithm & \bf Mean & \bf Max \\
		\midrule
		ERAC & \bf 28.30 $\pm$ 0.06 & \bf 28.42 \\
		ERAC w/o Future Ent. & 28.06 $\pm$ 0.05 & 28.11 \\
		AC & 28.04 $\pm$ 0.05 & 28.10 \\
		\bottomrule
	\end{tabular}
    \caption{Comparing ERAC with the variant without considering future entropy.}
    \label{tab:future_ent}
\end{table}
\section{Discussion}
In this work, motivated by the intriguing connection between the token-level RAML and the entropy-regularized RL, we propose two algorithms for neural sequence prediction.
Despite the distinct training procedures, both algorithms combine the idea of fine-grained credit assignment and the entropy regularization, leading to positive empirical results.

However, many problems remain widely open. 
In particular, the oracle Q-function $Q_\phi$ we obtain is far from perfect.
We believe the ground-truth reference contains sufficient information for such an oracle, and the current bottleneck lies in the RL algorithm.
Given the numerous potential applications of such an oracle, we believe improving its accuracy will be a promising future direction.

\clearpage
\bibliography{acl2018}
\bibliographystyle{acl_natbib}

\appendix
\onecolumn
\clearpage
\section{Proofs}
\label{sec:A-proof}

\subsection{Main Proofs}
\label{sec:A-main-proof}
\begin{customprop}{1}
For any ground-truth pair $(\mb{x}^*, \mb{y}^*)$,
$P_{Q_R}$ and $Q_R$ satisfy the following marginal match condition and terminal condition:

	\begin{equation}
		\prod_{t=1}^{|\mb{y}|} P_{Q_R}(y_t \mid \mb{y}_{1}^{t-1}) = P_R(\mb{y} \mid \mb{x}^* ) \quad \forall \mb{y} \in \mathcal{Y}
		\label{eqn:marg_app}
	\end{equation}
    
    \begin{equation}
    	Q_R(\hat{\mb{y}}, \eos; \mb{y}^*) = R(\hat{\mb{y}} + \eos; \mb{y}^*) - R(\hat{\mb{y}}; \mb{y}^*) \quad \forall \hat{\mb{y}} \in \mathcal{Y}^-
    	\label{eqn:ter_app}
    \end{equation}

if and only if for any $\mb{y} \in \mathcal{Y}$, 
	\begin{align}
	Q_R(\mb{y}_{1}^{t-1}, y_t; \mb{y}^*) = 
	\begin{cases}
	R(\mb{y}_{1}^t; \mb{y}^*)  - R(\mb{y}_{1}^{t-1}; \mb{y}^*) +\; \tau \log \sum_{w \in \mathcal{W}}\EXP{Q_R(\mb{y}_{1}^{t}, w; \mb{y}^*) / \tau}, & t < |\mb{y}| \\
	R(\mb{y}_{1}^t; \mb{y}^*)  - R(\mb{y}_{1}^{t-1}; \mb{y}^*), & t = |\mb{y}|
	\label{eqn:Q_cases_app}
	\end{cases}
	\end{align}

\end{customprop}

\begin{proof}
To avoid clutter, we drop the dependency on $\mb{x}^*$ and $\mb{y}^*$. 
The following proof holds for each possible pair of $(\mb{x}^*, \mb{y}^*)$. 

Firstly, it is easy to see that the terminal condition in Eqn. \eqref{eqn:ter_app} exactly corresponds to the $t=|\mb{y}|$ case of Eqn. \eqref{eqn:Q_cases_app}, since $y_t=\eos$ for $y\in \mathcal{Y}$.
So, we will focus on the non-terminal case next.

\paragraph{Sufficiency} For convenience, define $V_R(\mb{y}_{1}^{t}) = \tau \log \sum_{w \in \mathcal{W}} \EXP{Q_R(\mb{y}_{1}^{t}, w) / \tau}$. 
Suppose Eqn. \eqref{eqn:Q_cases_app} is true. Then for any $\mb{y} \in \mathcal{Y}$, 
\begin{align*}
P_{Q_R}(\mb{y})&=\prod_{t=1}^{|\mb{y}|} P_{Q_R}(y_t \mid \mb{y}_{1}^{t-1})\\
&=\EXP{ \frac{\sum_{t=1}^{|\mb{y}|} Q_R(\mb{y}_1^{t-1}, y_t)-V_R(\mb{y}_1^{t-1})}{\tau}}\\
&=\EXP{ \frac{\sum_{t=1}^{|\mb{y}|} \left[ R(\mb{y}_{1}^{t}) - R(\mb{y}_{1}^{t-1}) \right] + \sum_{t=1}^{|\mb{y}|-1}V_R(\mb{y}_1^{t}) -\sum_{t=1}^{|\mb{y}|}V_R(\mb{y}_1^{t-1})}{\tau}} \\
&=\EXP{ \frac{R(\mb{y})-V_R(\emptyset)}{\tau}} 
\end{align*}
where $V_R(\emptyset)$ denotes $V_R(\mb{y}_1^{t})$ when $t=0$ and $\mb{y}_1^{t}$ is an empty set. Since $P_{Q_R}(\mb{y})$ is a valid distribution by construction, we have 
\begin{equation*}
V_R(\emptyset)=\sum_{\mb{y}\in \mathcal{Y}}\EXP{ \frac{R(\mb{y})}{\tau}}
\end{equation*}
Hence, 
\begin{equation*}
P_{Q_R}(\mb{y}) = \frac{R(\mb{y})/\tau}{\sum_{\mb{y}'\in \mathcal{Y}} R(\mb{y}')/\tau}=P_R(\mb{y}),
\end{equation*}
which satisfies the marginal match requirement. 

\paragraph{Necessity}
Now, we show that the specific formulation of $Q_R$ (Eqn. \eqref{eqn:Q_cases_app}) is also a necessary condition of the marginal match condition (Eqn. \eqref{eqn:marg_app}).

The token-level target distribution can be simplified as
\begin{equation*}
P_{Q_R}(y_t \mid \mb{y}_{1}^{t-1})
= \frac{ \EXP{ Q_R(\mb{y}_{1}^{t-1}, y_t) / \tau } }{ \sum_{w \in \mathcal{W}} 
	\EXP{ Q_R(\mb{y}_{1}^{t-1}, w) / \tau } }=\EXP{\frac{Q_R(\mb{y}_{1}^{t-1}, y_t)-V_R(\mb{y}_1^{t-1})}{\tau}}.
\end{equation*}

Suppose Eqn. \eqref{eqn:marg_app} is true. 
For any $\mb{y} \in \mathcal{Y}^-$ and $t \leq |\mb{y}|$ and define $\mb{y}' = \mb{y}_1^t + \eos$ and $\mb{y}'' = \mb{y}_{1}^{t-1} + \eos$. Obviously, it follows $\mb{y}', \mb{y}'' \in \mathcal{Y}$.
Also, by definition,
\begin{align*}
P_R(\mb{y}') 
&= P_R(\eos \mid \mb{y}_1^{t}) \times P_R(y_t \mid \mb{y}_1^{t-1}) \times P_R(\mb{y}_1^{t-1}) \\
P_R(\mb{y}'') 
&= P_R(\eos \mid \mb{y}_1^{t-1}) \times P_R(\mb{y}_1^{t-1}) 
\end{align*}
Then, consider the ratio
\begin{align*}
\frac{P_R(\mb{y}')}{P_R(\mb{y}'')}
&= \frac{ P_R(\eos \mid \mb{y}_1^{t}) \times P_R(y_t \mid \mb{y}_1^{t-1}) \times \bcancel{P_R(\mb{y}_1^{t-1})} }
{ P_R(\eos \mid \mb{y}_1^{t-1}) \times \bcancel{P_R(\mb{y}_1^{t-1})} } \\
\EXP{\frac{R(\mb{y}') - R(\mb{y}'')}{\tau}} 
&= \EXP{\frac{Q_R(\mb{y}_1^{t}, \eos) - V_R(\mb{y}_1^{t})}{\tau}} \times 
\EXP{\frac{Q_R(\mb{y}_1^{t-1}, y_t) - \bcancel{V_R(\mb{y}_1^{t-1})} }{\tau}} \\
&\quad \Big/ \EXP{\frac{Q_R(\mb{y}_{1}^{t-1}, \eos) - \bcancel{V_R(\mb{y}_1^{t-1})} }{\tau}} \\
R(\mb{y}') - R(\mb{y}'')
&= Q_R(\mb{y}_1^t, \eos) - Q_R(\mb{y}_{1}^{t-1},\eos) - V_R(\mb{y}_1^{t}) + Q_R(\mb{y}_{1}^{t-1}, y_t).
\end{align*}
Now, by the terminal condition (Eqn. \eqref{eqn:ter_app}), we essentially have
\begin{align*}
Q_R(\mb{y}_1^t, \eos) &= R(\mb{y}_1^t + \eos) - R(\mb{y}_1^t) = 0 \\
Q_R(\mb{y}_1^{t-1}, \eos) &= R(\mb{y}_1^{t-1} + \eos) - R(\mb{y}_1^{t-1}) = 0 
\end{align*}
Thus, it follows
\begin{align*}
& R(\mb{y}') - R(\mb{y}'') = Q_R(\mb{y}_{1}^{t-1}, y_t) - V_R(\mb{y}_1^{t}) \\
\iff 
& Q_R(\mb{y}_{1}^{t-1}, y_t	) = R(\mb{y}_1^t) - R(\mb{y}_1^{t-1}) + \tau \log \sum_{w \in \mathcal{W}}\EXP{Q_R(\mb{y}_{1}^{t}, w) / \tau},
\end{align*}
which completes the proof.
\end{proof}

\begin{customcoro}{1}
	Please refer to \S\ref{sec:rl_equivalence} for the Corollary. 
\end{customcoro}

\begin{proof}
Similarly, we drop the dependency on $\mb{x}^*$ and $\mb{y}^*$ to avoid clutter. 
We first prove the equivalence of $Q^*(\mb{y}_1^{t-1}, y_t)$ with $Q_R(\mb{y}_1^{t-1}, y_t)$ by induction.

\begin{itemize}
	\item \textbf{Base case}: When $t = T$, for any $\mb{y} \in \mathcal{Y}$, $y_{T}$ can only be $\eos$. 
	So, by definition, we have 
	\begin{align*}
	&V^*(\mb{y}_1^{T-1}) = Q^*(\mb{y}_1^{T-1}, \eos) \\
	\iff
	&  \tau \log \sum_{a \in \mathcal{W}} \EXP{ Q^*(\mb{y}_1^{T-1}, a) / \tau } =  Q^*(\mb{y}_1^{T-1}, \eos) \\
	\implies
	& Q^*(\mb{y}_1^{T-1}, a) = -\infty, \forall a \neq \eos.
	\end{align*}
	Hence, 
	\begin{align*}
	Q^*(\mb{y}_1^{T-1}, y_{T}) &=
	\begin{cases}
	r(\mb{y}_1^{T-1}, \eos),& \text{if $y_{T} = \eos$} \\
	-\infty,& \text{otherwise}
	\end{cases}
	\end{align*}
	For the first case, it directly follows
	\[ 	Q^*(\mb{y}_1^{T-1}, \eos) = r(\mb{y}_1^{T-1}, \eos) = R(\mb{y}_1^{T-1} + \eos) - R(\mb{y}_1^{T-1}) = Q_R(\mb{y}_1^{T-1}, \eos).\]
	For the second case, since only $\eos$ is allowed to be generated, the target distribution $P_{Q_R}$ should be a single-point distribution at $\eos$. This is equivalent to define
	\begin{align*}
		Q_R(\mb{y}_1^{T-1}, a) = -\infty, \forall a \neq \eos,
	\end{align*}
	which proves the second case.
	Combining the two cases, it concludes
	\[ Q^*(\mb{y}_1^{T-1}, a) = Q_R(\mb{y}_1^{T-1}, a), \forall \mb{y} \in \mathcal{Y}, a \in \mathcal{W}. \]
	
	\item \textbf{Induction step}: When $0 < t < T$, assume the equivalence holds when $k > t$, i.e., 
	\[ Q^*(\mb{y}_1^{k-1}, w)=Q_R(\mb{y}_1^{k-1}, w), \forall k > t, w \in \mathcal{W}.\] 
	Then, 
	\begin{align*}
	Q^*(\mb{y}_1^{t-1}, y_t) 
	&= r(\mb{y}_1^{t-1}, y_t) + \gamma \E_{s' \sim \rho_s} [ \alpha \log \sum_{a \in \mathcal{A}} \EXP{ Q^*(s', a) / \alpha }] \\
	&= r(\mb{y}_1^{t-1}, y_t) + \tau \log \sum_{a \in \mathcal{W}} \EXP{ Q^*(\mb{y}_1^{t}, a) / \tau } & (\alpha = \tau, \mathcal{A} = \mathcal{W}) \\
	&=r(\mb{y}_1^{t-1}, y_t) + \tau \log \sum_{a \in \mathcal{W}} \EXP{ Q_R(\mb{y}_1^{t}, a) / \tau } & (Q^*(\mb{y}_1^{k}, a)=Q_R(\mb{y}_1^{k}, a) \mbox{ for } k \geq t) \\
	&=Q_R(\mb{y}_1^{t-1}, y_t). 
	\end{align*}
\end{itemize}
Thus, $Q^*(\mb{y}_1^{t-1}, y_t)=Q_R(\mb{y}_1^{t-1}, y_t)$ holds for $t\in [1, T]$.

With the equivalence between $Q_R$ and $Q^*$, we can easily prove $V^*=V_R$ and $\pi^*=P_{Q_R}$, 
\begin{align*}
V^*(\mb{y}_1^{t-1}) &= \alpha \log \sum_{a \in \mathcal{A}} \EXP{ Q^*(\mb{y}_1^{t-1}, a) / \alpha } \\
&=\tau \log \sum_{a \in \mathcal{W}} \EXP{ Q^*(\mb{y}_1^{t-1}, a) / \tau } & (\alpha = \tau, \mathcal{A} = \mathcal{W}) \\
&=V_R(\mb{y}_1^{t-1}) \\
\pi^*(y_t \mid \mb{y}_1^{t-1}) 
&= \frac{ \EXP{ Q^*(\mb{y}_1^{t-1}, y_t) / \tau } }{ \sum_{w \in \mathcal{W}} \EXP{ Q^*(\mb{y}_1^{t-1}, y_t) / \tau } } \\
&= \frac{ \EXP{ Q_R(\mb{y}_1^{t-1}, y_t) / \tau } }{ \sum_{w \in \mathcal{W}} \EXP{ Q_R(\mb{y}_1^{t-1}, y_t) / \tau } } \\
&=P_{Q_R}(y_t \mid \mb{y}_1^{t-1}) 
\end{align*}

\end{proof}

\subsection{Other Proofs}
\label{sec:A-other-proof}
We derive the equivalence between the VAML's objective (Eqn. \eqref{eqn:vaml_objective}) and the RAML's objective (Eqn. \eqref{eqn:raml_objective}).
\begin{align*}
& \CE{P_{Q_\phi}}{P_\theta} \\
=& -\E_{\mb{y} \sim P_{Q_\phi}} \log P_\theta(\mb{y}) \\
=& -\E_{\mb{y} \sim P_{Q_\phi}} \sum_{t=1}^{|\mb{y}|} \log P_\theta(y_t \mid \mb{y}_1^{t-1}) \\
=& -\sum_{t=1}^{T} \E_{\mb{y}_1^t \sim P_{Q_\phi}(Y_1^{t})} \log P_\theta(y_t \mid \mb{y}_1^{t-1}) & \text{($T$ is longest possible length)} \\
=& \sum_{t=1}^{T} \E_{\mb{y}_1^{t-1} \sim P_{Q_\phi}(\mb{Y}_1^{t-1})} \left[ -\E_{y_t \sim P_{Q_\phi}(Y_t \mid \mb{y}_1^{t-1})} \log P_\theta(y_t \mid \mb{y}_1^{t-1}) \right] \\
=& \sum_{t=1}^{T} \E_{\mb{y}_1^{t-1} \sim P_{Q_\phi}(\mb{Y}_1^{t-1})} \CE{P_{Q_\phi}(Y_t \mid \mb{y}_1^{t-1})}{P_\theta(Y_t \mid \mb{y}_1^{t-1})} \\
=& \sum_{t=1}^{T} \E_{\mb{y}_1^{t-1} \sim P_{Q_\phi}(\mb{Y}_1^{t-1})} \sum_{y_t \in \mathcal{W}} P_{Q_\phi}(y_t \mid \mb{y}_1^{t-1}) \underbrace{\CE{P_{Q_\phi}(Y_t \mid \mb{y}_1^{t-1})}{P_\theta(Y_t \mid \mb{y}_1^{t-1})}}_\text{const. w.r.t. $y_t$} \\
=& \sum_{t=1}^{T} \underbrace{\E_{\mb{y}_1^{t-1} \sim P_{Q_\phi}(\mb{Y}_1^{t-1})} \E_{y_t \in P_{Q_\phi}(W \mid \mb{y}_1^{t-1})}}_{\E_{\mb{y}_1^{t} \sim P_{Q_\phi}(\mb{Y}_1^{t})}}
	 \left[ \CE{P_{Q_\phi}(Y_t \mid \mb{y}_1^{t-1})}{P_\theta(Y_t \mid \mb{y}_1^{t-1})} \right] \\
=& \sum_{t=1}^{T} \E_{\mb{y}_1^{t} \sim P_{Q_\phi}(\mb{Y}_1^{t})} \left[ \CE{P_{Q_\phi}(Y_t \mid \mb{y}_1^{t-1})}{P_\theta(Y_t \mid \mb{y}_1^{t-1})} \right] \\
=& \E_{\mb{y} \sim P_{Q_\phi}(\mb{Y})}  \sum_{t=1}^{|\mb{y}|} \CE{P_{Q_\phi}(Y_t \mid \mb{y}_1^{t-1})}{P_\theta(Y_t \mid \mb{y}_1^{t-1})} 
\end{align*}

\section{Implementation Details}
\label{sec:A-implementation}

\subsection{RAML}
\label{sec:A-raml}
In RAML, we want to optimize the cross entropy $\CE{P_R(\mb{Y} \mid \mb{x}^*, \mb{y}^*)}{P_\theta(\mb{Y} \mid \mb{x}^*)}$.
As discussed in \S\ref{sec:raml}, directly sampling from the exponentiated pay-off distribution $P_R(Y\mid x^*)$ is impractical.
Hence, normalized importance sampling has been exploited in previous work~\citep{norouzi2016reward,ma2017softmax}.
Define the proposal distribution to be $P_S(\mb{Y} \mid \mb{x}^*, \mb{y}^*)$. Then, the objective can be rewritten as
\begin{align*}
\CE{P_R(\mb{Y} \mid \mb{x}^*, \mb{y}^*)}{P_\theta(\mb{Y} \mid \mb{x}^*)}
&= -\E_{\mb{y} \sim P_S(\mb{Y} \mid \mb{x}^*, \mb{y}^*)} \frac{P_R(\mb{y} \mid \mb{x}^*, \mb{y}^*)}{P_S(\mb{y} \mid \mb{x}^*, \mb{y}^*)} \log P_{\theta}(\mb{y} \mid \mb{x}^*) \\
&= -\E_{\mb{y} \sim P_S(\mb{Y} \mid \mb{x}^*, \mb{y}^*)} 
\frac{ \frac{ \EXP{R(\mb{y}, \mb{y}^*) / \tau} }{ \tilde{P}_S(\mb{y} \mid \mb{x}^*, \mb{y}^*) } }{ 
	\E_{\mb{y}' \sim P_S(\mb{Y} \mid \mb{x}^*, \mb{y}^*)} \frac{ \EXP{R(\mb{y}', \mb{y}^*) / \tau} }{ \tilde{P}_S(\mb{y}' \mid \mb{x}^*, \mb{y}^*)} } 
\log P_{\theta}(\mb{y} \mid \mb{x}^*) \\
&= -\E_{\mb{y} \sim P_S(\mb{Y} \mid \mb{x}^*, \mb{y}^*)} 
\frac{ w(\mb{y}, \mb{y}^*) }{ \E_{\mb{y}' \sim P_S(\mb{Y} \mid \mb{x}^*, \mb{y}^*)} w(\mb{y}', \mb{y}^*) }  \log P_{\theta}(\mb{y} \mid \mb{x}^*) \\
&\approx -\sum_{i=1}^{M}
\frac{ w(\mb{y}^{(i)}, \mb{y}^*) }{\sum_{i=1}^{M} w(\mb{y}^{(i)}, \mb{y}^*) }  \log P_{\theta}(\mb{y}^{(i)} \mid \mb{x}^*),
\end{align*}
where $w(\mb{y}, \mb{y}^*) = \frac{ \EXP{R(\mb{y}, \mb{y}^*) / \tau} }{ \tilde{P}_S(\mb{y} \mid \mb{x}^*, \mb{y}^*) }$ is the unnormalized importance weight, $\tilde{P}_S$ denotes the unnormalized probability of $P_S = \frac{\tilde{P}_S}{Z}$, $M$ is the number of samples used, and $\mb{y}^{(i)}$ is the $i$-th sample drawn from the proposal distribution $P_S(\mb{Y} \mid \mb{x}^*, \mb{y}^*)$.

With importance sampling, the problem turns to what proposal distribution we should use. 
In the original work~\citep{norouzi2016reward}, the proposal distribution is defined by the hamming distance as used. \citet{ma2017softmax} find that it suffices to perform $N$-gram replacement of the reference sentence. Specifically, $P_S(\mb{Y} \mid \mb{x}^*, \mb{y}^*)$ can be a uniform distribution defined on set $\mathcal{Y}_\text{ngram}$ where $\mathcal{Y}_\text{ngram}$ is obtained by randomly replacing an $n$-gram of $\mb{y}^*$ ($n\leq 4$).

In this work, we adapt the simple $n$-gram replacement distribution, denoted as $P_\text{ngram}(\mb{Y} \mid \mb{x}^*, \mb{y}^*)$, which simplifies the RAML objective into
\begin{equation*}
\min_{\theta} -\sum_{i=1}^{M}
\frac{ \EXP{R(\mb{y}^{(i)}, \mb{y}^*) / \tau} }{\sum_{i=1}^{M} \EXP{R(\mb{y}^{(i)}, \mb{y}^*) / \tau} }  \log P_{\theta}(\mb{y}^{(i)} \mid \mb{x}^*) 
\end{equation*}
Following \citet{ma2017softmax}, we make sure the reference sequence is always among the $M$ samples used.

\subsection{VAML}
\label{sec:A-vaml}
As discussed in \S\ref{sec:algo}, the VAML training consists of two phases:
\begin{itemize}[leftmargin=1.0em]
\item In the first phase, Soft Q-Learning is used to train $Q_\phi$ based on Eqn. \eqref{eqn:soft_q_learning}.
Since Soft Q-Learning accepts off-policy trajectories, in this work, we use two types of off-policy sequences:
\begin{enumerate}
	\item The first type is simply the ground-truth sequence, which provides strong learning signals.
	\item The second type of sequences is actually drawn from the same $n$-gram replacement distribution discussed above. The reason is that in the second training phase, such $n$-gram replaced trajectories will be used. Since the learned $Q_\phi$ won't be perfect, we hope the exposing $Q_\phi$ with these trajectories can improve its accuracy on them, making the second phase of training easier.
\end{enumerate}
Algorithm \ref{algo:soft_q_learning} summarizes the first phase.
\begin{algorithm}
	\caption{VAML Phase 1: Soft Q-Learning to approximate $Q^*$}
	\label{algo:soft_q_learning}
	\begin{algorithmic}[1]
		\REQUIRE 
		A Q-function approximator $Q_{\phi}$ with parameter $\phi$, and the hyper-parameters $\tau$, $M$. 
		\WHILE{Not Converged}
		\STATE Receive a random example $(\mb{x}^*, \mb{y}^*)$.
		\STATE Sample $M-1$ sequences $\{\mb{y}^{(i)}\}_{i=1}^{M-1}$ from $P_\text{ngram}(\mb{Y} \mid \mb{x}^*, \mb{y}^*)$ and let $\mb{y}^{(M)} = \mb{y}^*$.
		\STATE Compute all the rewards $r(\mb{y}_{1}^{t-1}, y_t; \mb{y}^*)$ for each $\mb{y} \in \{\mb{y}^{(i)}\}_{i=1}^{M}$ and $t = 1, \dots, |\mb{y}|$.
		\STATE Compute the target Q-values for each $\mb{y} \in \{\mb{y}^{(i)}\}_{i=1}^{M}$  and $t = 1, \dots, |\mb{y}|$
		\[ \hat{Q}_\phi(\mb{y}_{1}^{t-1}, y_t; \mb{y}^*) = r(\mb{y}_{1}^{t-1}, y_t; \mb{y}^*) + \tau \log \sum_{w \in \mathcal{W}} \EXP{ Q_\phi(\mb{y}_{1}^{t}, w; \mb{y}^*) / \tau}. \]
		\STATE Compute the Soft-Q Learning loss 
		\[ \mathcal{L}_\text{SoftQ} = \frac{1}{M} \sum_{i=1}^{M} \sum_{t=1}^{|\mb{y}^{(i)}|} \left\| Q_\phi({\mb{y}^{(i)}}_{1}^{t-1}, y_t^{(i)}; \mb{y}^*) - \hat{Q}_\phi({\mb{y}^{(i)}}_{1}^{t-1}, y_t^{(i)}; \mb{y}^*) \right\|_2^2. \]
		\STATE Update $Q_\phi$ according to the loss $\mathcal{L}_\text{SoftQ}$. 
		\ENDWHILE
	\end{algorithmic}
\end{algorithm}

\item Once the $Q_\phi$ is well trained in the first phase, the second phase is to minimize the cross entropy $\CE{P_{Q_\phi}(\mb{Y} \mid \mb{x}^*, \mb{y}^*)}{P_\theta(\mb{Y} \mid \mb{x}^*)}$ based on Eqn. \eqref{eqn:vaml_objective}, i.e.,
\begin{align*}
\min_\theta \E_{\mb{y} \sim P_{Q_\phi}} \left[ \sum_{t=1}^{|\mb{y}|} \CE{P_{Q_\phi}(Y_t \mid \mb{y}_1^{t-1})}{ P_{\theta}(Y_t \mid \mb{y}_1^{t-1}) } \right].
\end{align*}
Ideally, we would like to directly sample from $P_{Q_\phi}$, and perform the optimization. However, we find samples from $P_{Q_\phi}$ are quite similar to each other. We conjecture this results from both the imperfect training in the first phase, and the intrinsic difficulty of getting diverse samples from an exponentially large space when the distribution is high concentrated.

Nevertheless, for this work, we fall back to the same importance sampling method as used in RAML and use the $n$-gram replacement distribution as the proposal. Hence, the objective becomes
\begin{align*}
  & \E_{\mb{y} \sim P_{Q_\phi}} \left[ \sum_{t=1}^{|\mb{y}|} \CE{P_{Q_\phi}(Y_t \mid \mb{y}_1^{t-1})}{ P_{\theta}(Y_t \mid \mb{y}_1^{t-1}) } \right] \\
=& \E_{\mb{y} \sim P_\text{ngram}} \left[ \frac{ w(\mb{y}, \mb{y}^*) }{ \E_{\mb{y}' \sim P_\text{ngram}(\mb{Y} \mid \mb{x}^*, \mb{y}^*)} w(\mb{y}', \mb{y}^*) }  \sum_{t=1}^{|\mb{y}|} \CE{P_{Q_\phi}(Y_t \mid \mb{y}_1^{t-1})}{ P_{\theta}(Y_t \mid \mb{y}_1^{t-1}) } \right] \\
\approx& \sum_{i=1}^{M}
\frac{ \EXP{R(\mb{y}^{(i)}, \mb{y}^*) / \tau} }{ \sum_{i=1}^{M} \EXP{R(\mb{y}^{(i)}, \mb{y}^*) / \tau} }
\left[ \sum_{t=1}^{|\mb{y}^{(i)}|} \CE{P_{Q_\phi}(Y_t \mid {\mb{y}^{(i)}}_1^{t-1})}{ P_{\theta}(Y_t \mid {\mb{y}^{(i)}}_1^{t-1}) } \right].
\end{align*}
However, we found directly using this objective does not yield improved performance compared to RAML, mostly likely due to some erratic estimations of $Q_\phi$.
Thus, we only use this objective for some step with certain probability $\kappa \in (0, 1)$, leaving others trained by MLE. 
Formally, define 
\begin{equation*}
\mathcal{J}_\kappa(\mb{y}_1^t) = \E_{z \sim \mathrm{Bernoulli}(\kappa)} \left[z \CE{P_{Q_\phi}(Y_t \mid \mb{y}_1^{t-1})}{ P_{\theta}(Y_t \mid \mb{y}_1^{t-1}) } - (1 - z) \log P_{\theta}(y_t \mid \mb{y}_1^{t-1})\right] ,
\end{equation*} 
the VAML objective practically used is
\begin{equation*}
\min_{\theta} 
\sum_{i=1}^{M}
\frac{ \EXP{R(\mb{y}^{(i)}, \mb{y}^*) / \tau} }{ \sum_{i=1}^{M} \EXP{R(\mb{y}^{(i)}, \mb{y}^*) / \tau}}
\left[ \sum_{t=1}^{|\mb{y}^{(i)}|} \mathcal{J}_\kappa({\mb{y}^{(i)}}_1^t) \right].
\end{equation*}
Algorithm \ref{algo:vaml} summarizes the second phase.
\begin{algorithm}
	\caption{VAML Phase 2: Sequence model training with token-level target}
	\label{algo:vaml}
	\begin{algorithmic}[1]
		\REQUIRE 
		A sequence prediction model $P_{\theta}$ with parameter $\theta$, 
		a pre-trained Q-function approximator $Q_{\phi}$, 
		and hyper-parameters $\tau$, $M$, $\kappa$
		\WHILE{Not Converged}
		\STATE Receive a random example $(\mb{x}^*, \mb{y}^*)$.
		\STATE Sample $M-1$ sequences $\{\mb{y}^{(i)}\}_{i=1}^{M-1}$ from $P_\text{ngram}(\mb{Y} \mid \mb{x}^*, \mb{y}^*)$ and let $\mb{y}^{(M)} = \mb{y}^*$.
		\STATE Compute the VAML loss using
		\[ \mathcal{L}_\text{VAML} = \sum_{i=1}^{M} \frac{ \EXP{R(\mb{y}^{(i)}, \mb{y}^*) / \tau} }{ \sum_{i=1}^{M} \EXP{R(\mb{y}^{(i)}, \mb{y}^*) / \tau} } \left[ \sum_{t=1}^{|\mb{y}^{(i)}|} \mathcal{J}_\kappa({\mb{y}^{(i)}}_1^t) \right]. \]
		\STATE Update $P_\theta$ according to the loss $\mathcal{L}_\text{VAML}$. 
		\ENDWHILE
	\end{algorithmic}
\end{algorithm}
\end{itemize}

\subsection{ERAC}
\label{sec:A-erac-algo}
Following ~\citet{bahdanau2016actor}, we first pre-train the actor, then train the critic with the fixed actor and finally fine-tune them together. The specific procedure for training ERAC is 
\begin{itemize}[topsep=0.5em,itemsep=0em]
	\item Pre-training the actor using maximum likelihood training
	\item Pre-training the critic using Algorithm \ref{algo:ERAC} with the actor fixed
	\item Fine-tuning both the actor and critic with Algorithm \ref{algo:ERAC}
\end{itemize}

\begin{algorithm}[t]
    \caption{ERAC Algorithm}
    \label{algo:ERAC}
    \begin{algorithmic}[1]
    \REQUIRE 
        A critic $Q_{\phi}(\mb{y}_1^{t-1}, y_t; \mb{y}^*)$
        and an actor $\pi_{\theta}(w \mid \mb{y}_1^{t})$ with
        weights $\phi$ and $\theta$ respectively, 
        and hyper-parameters $\tau$, $\beta$, $\lambda_\text{var}$, $\lambda_\text{mle}$
    \STATE Initialize delayed target critic $Q_{\bar{\phi}}$ with the same weights: $\bar{\phi} = \phi$.
    \WHILE{Not Converged}
    \STATE Receive a random example $(\mb{x}^*, \mb{y}^*)$.
    \STATE Generate a sequence $\mb{y}$ from $\pi_{\theta}$.
    \STATE Compute the rewards $r(\mb{y}_{1}^{t-1}, y_t; \mb{y}^*)$ for $t = 1, \dots, |\mb{y}|$.
    \STATE Compute targets for the critic
    \begin{equation*}
    \hat{Q}_{\bar{\phi}}(\mb{y}_{1}^{t-1}, y_t; \mb{y}^*) 
	= r(\mb{y}_{1}^{t-1}, y_t) + \tau \, \mathcal{H}(\pi_\theta(\cdot \mid \mb{y}_1^t)) 
	+ \sum_{w \in \mathcal{W}} \pi_\theta(w \mid \mb{y}_1^t) Q_{\bar{\phi}}(\mb{y}_1^t, w; \mb{y}^*).    
	\end{equation*}       
    \STATE Compute loss for critic
    \begin{align*}
    \mathcal{L}_\text{critic}
    &=\sum_{t=1}^{|\mb{y}|} \left[ Q_\phi(\mb{y}_{1}^{t-1}, y_t; \mb{y}^*) - \hat{Q}_{\bar{\phi}}(\mb{y}_{1}^{t-1}, y_t; \mb{y}^*) \right]^2 + \lambda_\text{var} \sum_{w \in \mathcal{W}} \left[ Q_\phi(\mb{y}_{1}^{t-1}, w; \mb{y}^*) - \bar{Q}_\phi(\mb{y}_{1}^{t-1}; \mb{y}^*) \right]^2, \\
    &\text{where}\quad \bar{Q}_\phi(\mb{y}_{1}^{t-1}; \mb{y}^*)=\frac{1}{|\mathcal{W}|} \sum_{w' \in \mathcal{W}} Q_\phi(\mb{y}_{1}^{t-1}, w'; \mb{y}^*)
    \end{align*}
    \STATE Compute loss for actor
    \begin{align*}
    \mathcal{L}_\text{actor} &= -\left[ \sum_{t=1}^{|\mb{y}|}  \sum_{w \in \mathcal{W}} \pi_{\theta}(w \mid \mb{y}_1^{t-1}) Q_\phi(\mb{y}_1^{t-1}, w; \mb{y}^*) \nonumber + \tau \mathcal{H}(\pi_{\theta}(\cdot \mid \mb{y}_1^{t-1})) + \lambda_\text{mle}\sum_{t=1}^{|\mb{y}^*|}\log \pi_{\theta}(y_t^* \mid {\mb{y}^*}_1^{t-1})\right]
    \end{align*}
    \STATE Update critic according to the loss $\mathcal{L}_\text{critic}$. 
    \STATE If actor is not fixed, update actor according to the loss $\mathcal{L}_\text{actor}$
    \STATE Update delayed target critic: $\bar{\phi} = \beta \phi + (1 - \beta) \bar{\phi}$
    \ENDWHILE
    \end{algorithmic}
\end{algorithm}

\subsection{Hyper-parameters}
\label{sec:A-hyperparam}
\paragraph{RAML \& VAML}
The hyper-parameters for RAML and VAML training are summarized in Tab. \ref{tab:raml_vaml}. 
We set the gradient clipping value to $5.0$ for both the Q-function approximator $Q_{\phi}$ and the sequence prediction model $P_{\theta}$, except for the sequence prediction model in the captioning task where the gradient clipping value is set to $1.0$. 
{\fontsize{9}{11}
\begin{table*}[!h]
	\centering
	\begin{tabular}{l|ccc|ccc}
		\toprule
		& \multicolumn{3}{c|}{\bf Machine Translation} & \multicolumn{3}{c}{\bf Image Captioning} \\
		\bf Hyper-parameters & VAML-1 & VAML-2 & RAML & VAML-1 & VAML-2 & RAML  \\
		\midrule
		optimizer & Adam & SGD & SGD & Adam & SGD & SGD \\
		learning rate & 0.001 & 0.6 & 0.6 & 0.001 & 0.5 & 0.5 \\
		batch size & 50 & 42 & 42 & 32 $\times$ 5 & 32 $\times$ 5 & 32 $\times$ 5\\
		$M$ & 5 & 5 & 5 & 2 & 6 & 6\\
		$\tau$ & 0.4 & 0.4 & 0.4 & 0.7 & 0.7 & 0.7 \\
		$\kappa$ & N.A. & 0.2 & N.A. & N.A. & 0.1 & N.A. \\
		\bottomrule
	\end{tabular}
	\caption{Optimization related hyper-parameters of RAML and VAML for two tasks. ``VAML-1'' and ``VAML-2'' indicate the phase 1 and phase 2 of VAML training respectively. ``N.A.'' means not applicable. ``32 $\times$ 5'' indicates using 32 images each with 5 reference captions.}
	\label{tab:raml_vaml}
\end{table*}}

\paragraph{AC \& ERAC} As described in \S\ref{sec:A-erac-algo}, the training using AC and ERAC involves three phases.
The hyper-parameters used for ERAC training in each phase are summarized in Table \ref{tab:hyperparam-erac}.
In all phases, the learning rate is halved when there is no improvement on the validation set. 
We use the same hyper-parameters for AC training, except that the entropy regularization coefficient $\tau$ is 0. 
Similar to the VAML case, the gradient clipping value is set to $5.0$ for both the actor and the critic, except that we set the gradient clipping value to $1.0$ for the actor in the captioning task.
\begin{table*}[!h]
	\centering
	\begin{tabular}{l|c|c|c}
		\toprule
		\bf Hyper-parameters & \bf MT w/ input feeding & \bf MT w/o input feeding & \bf Image Captioning \\
		\midrule
		\multicolumn{4}{c}{\bf Pre-train Actor} \\
		\midrule
		optimizer & SGD & SGD & SGD \\
		learning rate & 0.6 & 0.6 & 0.5 \\
		batch size & 50 & 50 & 32 $\times$ 5 \\
		\midrule
		\multicolumn{4}{c}{\bf Pre-train Critic} \\
		\midrule
		optimizer                                        & Adam & Adam & Adam \\
		learning rate                                   & 0.001 & 0.001 & 0.001 \\
		batch size                                       & 50      & 50     & 32 $\times$ 5 \\
		$\tau$ (entropy regularization)       & 0.05   & 0.04  & 0.01 \\
		$\beta$ (target net speed)              & 0.001 & 0.001 & 0.001 \\
		$\lambda_\text{var}$ (smoothness) & 0.001 & 0.001 & 0.001 \\
		\midrule
		\multicolumn{4}{c}{\bf Joint Training} \\
		\midrule
		optimizer                                        & Adam   & Adam  & Adam \\
		learning rate                                   & 0.0001 & 0.0001 & 0.0001 \\
		batch size                                       & 50       & 50       & 32 $\times$ 5 \\
		$\tau$ (entropy regularization)       & 0.05    & 0.04    & 0.01 \\
		$\beta$ (target net speed)              & 0.001   & 0.001  & 0.001 \\	
		$\lambda_\text{var}$ (smoothness) & 0.001  & 0.001   & 0.001 \\
		$\lambda_\text{MLE}$                     & 0.1       & 0.1       & 0.1 \\
		\bottomrule
	\end{tabular}
	\caption{Hyper-parameters for ERAC training}
	\label{tab:hyperparam-erac}
\end{table*}

\newpage
\section{Comparison with Previous Work}
\label{sec:A-comparison}
The detailed comparison with previous work in shown in Table \ref{tab:comparison_detailed}.
Under different comparable architectures (1 layer or 2 layers), ERAC outperforms previous algorithms with a clear margin.
\begin{table}[!h]
	\begin{adjustwidth}{-.5in}{-.5in}
	\captionsetup{font=small}
	\centering
	\small
	\begin{tabular}{l|cc|cccc|c}
		\toprule
		\bf \multirow{2}{*}{Algorithm} & \multicolumn{2}{c|}{\bf Encoder} & \multicolumn{4}{c|}{\bf Decoder} & \bf \multirow{2}{*}{BLEU } \\
& NN Type & Size & NN Type & Size & Attention & Input Feed & \\
		\midrule
		MIXER~\cite{ranzato2015sequence} & 1-layer CNN & 256                      & 1-layer LSTM & 256 & Dot-Prod & N & 20.73 \\
		BSO~\cite{wiseman2016sequence}   & 1-layer BiLSTM & 128 $\times$ 2 & 1-layer LSTM & 256 & Dot-Prod & Y & 27.9 \\
		Q(BLEU)~\cite{li2017learning}          & 1-layer BiLSTM & 128 $\times$ 2 & 1-layer LSTM & 256 & Dot-Prod & Y & 28.3 \\
		AC~\cite{bahdanau2016actor}         & 1-layer BiGRU & 256 $\times$ 2   & 1-layer GRU & 256 & MLP & Y & 28.53 \\
		RAML~\cite{ma2017softmax}           & 1-layer BiLSTM & 256 $\times$ 2 & 1-layer LSTM & 256 & Dot-Prod & Y & 28.77 \\
		\midrule 
		VAML          & 1-layer BiLSTM & 128 $\times$ 2 & 1-layer LSTM & 256 & Dot-Prod & Y & 28.94 \\
		ERAC          & 1-layer BiLSTM & 128 $\times$ 2 & 1-layer LSTM & 256 & Dot-Prod & Y & \bf 29.36 \\
		\midrule\midrule
		NPMT~\cite{huang2017toward} & 2-layer BiGRU & 256 $\times$ 2            & 2-layer LSTM & 512 & N.A. & N.A. & 29.92 \\
		NPMT+LM~\cite{huang2017toward} & 2-layer BiGRU & 256 $\times$ 2      & 2-layer LSTM & 512 & N.A. & N.A. & 30.08 \\
		\midrule
		ERAC          & 2-layer BiLSTM & 256 $\times$ 2 & 2-layer LSTM & 512 & Dot-Prod & Y & \bf 30.85 \\
		\bottomrule
	\end{tabular}
    \caption{Comparison of algorithms with detailed architecture information on the IWSTL 2014 dataset for MT.}
    \label{tab:comparison_detailed}
    \end{adjustwidth}
\end{table}

\end{document}